\documentclass[letterpaper]{article} 
\usepackage{aaai24}  
\usepackage{times}  
\usepackage{helvet}  
\usepackage{courier}  
\usepackage[hyphens]{url}  
\usepackage{graphicx} 
\usepackage{natbib}  
\usepackage{caption} 
\usepackage{algorithm}
\usepackage{algorithmic}

\usepackage{newfloat}
\usepackage{listings}
\DeclareCaptionStyle{ruled}{labelfont=normalfont,labelsep=colon,strut=off} 
\floatstyle{ruled}
\newfloat{listing}{tb}{lst}{}
\floatname{listing}{Listing}
\usepackage{bm}
\usepackage{amssymb}
\usepackage{amsthm}
\usepackage{mathrsfs}
\usepackage{indentfirst}
\usepackage{multirow}
\usepackage{subfig}
\usepackage{xcolor}
\nocopyright
\title{Data-Free Generalized Zero-Shot Learning}
\author {
    Bowen Tang\textsuperscript{\rm 1}\equalcontrib,
    Long Yan\textsuperscript{\rm 1}\equalcontrib,
    Jing Zhang\textsuperscript{\rm 1}\thanks{Corresponding author.}\equalcontrib,
    Qian Yu\textsuperscript{\rm 1},
    Lu Sheng\textsuperscript{\rm 1},
    Dong Xu\textsuperscript{\rm 2}
}
\affiliations {
    \textsuperscript{\rm 1} School of Software, Beihang University\\
    \textsuperscript{\rm 2} Department of Computer Science, The University of Hong Kong\\
    tbw.broHan@qq.com, \{yanlong,zhang\underline{ }jing, qianyu, lsheng\}@buaa.edu.cn, dongxu@hku.hk

}

\usepackage{bibentry}

\begin{document}

\maketitle

\begin{abstract}
Deep learning models have the ability to extract rich knowledge from large-scale datasets. However, the sharing of data has become increasingly challenging due to concerns regarding data copyright and privacy. Consequently, this hampers the effective transfer of knowledge from existing data to novel downstream tasks and concepts. Zero-shot learning (ZSL) approaches aim to recognize new classes by transferring semantic knowledge learned from base classes. However, traditional generative ZSL methods often require access to real images from base classes and rely on manually annotated attributes, which presents challenges in terms of data restrictions and model scalability. To this end, this paper tackles a challenging and practical problem dubbed as data-free zero-shot learning (DFZSL), where only the CLIP-based base classes data pre-trained classifier is available for zero-shot classification. Specifically, we propose a generic framework for DFZSL, which consists of three main components. Firstly, to recover the virtual features of the base data, we model the CLIP features of base class images as samples from a von Mises-Fisher (vMF) distribution based on the pre-trained classifier. Secondly, we leverage the text features of CLIP as low-cost semantic information and propose a feature-language prompt tuning (FLPT) method to further align the virtual image features and textual features. Thirdly, we train a conditional generative model using the well-aligned virtual image features and corresponding semantic text features, enabling the generation of new classes features and achieve better zero-shot generalization. Our framework has been evaluated on five commonly used benchmarks for generalized ZSL, as well as 11 benchmarks for the base-to-new ZSL. The results demonstrate the superiority and effectiveness of our approach. Our code is available in https://github.com/ylong4/DFZSL.

\end{abstract}

\section{Introduction}
The power of deep learning models lies in their ability to extract rich knowledge, including visual features and semantic information, from large-scale datasets. However, the sharing of data across different companies, institutions, and countries has become increasingly challenging and sensitive. Concerns related to data copyright and privacy, particularly in sensitive domains such as health and security, pose significant obstacles to the seamless transfer of knowledge from large-scale datasets to novel downstream tasks and concepts. These challenges impede the widespread utilization of deep learning models and limit their potential impact in various fields.

Inspired by the mounting concerns regarding data and model privacy issues, particularly in the context of knowledge transfer to new concepts, this paper addresses the problem of data-free zero-shot learning without access to any real data. Zero-shot learning (ZSL) addresses the challenge of recognizing new classes by leveraging semantic knowledge transferred from base classes. 
Despite the notable advancements in ZSL, 
most ZSL methods often require access to labeled images from base classes, either for aligning visual-semantic embeddings or training conditional generative models~\cite{s41:AWA2,f-CLSWGAN,TF-VAEGAN}. Unfortunately, obtaining real data from base classes is often impractical in real-world applications due to privacy or copyright restrictions. Moreover, existing approaches heavily rely on manually annotated attributes, which present challenges in terms of scalability and the difficulty of annotation~\cite{APY,t15,t17,t18}.  

The recent progress in large-scale pre-trained vision-language models, such as CLIP~\cite{CLIP}, have demonstrated impressive zero-shot generalization abilities. These models achieve this capability through extensive training on vast collections of image-caption pairs without the requirement of manually annotated attributes. However, effectively transferring the knowledge from these models, which are trained on weakly aligned image-caption pairs, to downstream fine-grained zero-shot classification tasks remains challenging and sub-optimal. This is primarily due to the discrepancy in class granularity between the pre-trained models and the specific classification tasks at hand. Propmt tuning deals with this issue by adding learnable prompts to the inputs. However the recent prompt tuning methods~\cite{CoOp,VP,Co-CoOp} still suffer from single-side alignment and rely on the access to the real images.

To this end, this paper addresses a challenging and practical problem dubbed as data-free zero-shot learning (DFZSL). 
In this setting, the only available resource for zero-shot classification is a pre-trained base classes classifier based on CLIP features. Notably, we do not have access to any real data from either the base or new classes, and manual attribute annotations are not required. Our setting is closely related to Absolute Zero-Shot Learning~\cite{t7}. However, their method still relies on manual attribute annotations and performs poorly in both conventional and generalized ZSL.

The proposed framework consists of three main components. Firstly, to recover the base class data, we model the CLIP features of base class images as samples from a von Mises-Fisher (vMF) distribution, with learnable mean ($\bm{\mu}$) and proper concentration ($\kappa$) parameters based on the pre-trained classifier. This allows us to recover the virtual features of the base data by sampling from the distribution. It is important to note that our method does not recover the original images. Instead, our focus is on recovering the high-level image feature vectors, which is more efficient and avoids the privacy and copyright concerns. Secondly, to bridge the base and new classes, we leverage the text encoder of CLIP to obtain low-cost semantic information in the form of generalizable text features, which eliminates the need for manual attribute annotations. Our framework is generic, and any vision-language foundation models can be potentially used. 
In order to enhance the adaptation to downstream fine-grained zero-shot classification tasks, we introduce a feature-language prompt tuning method. This method aims to further align the virtual image features of base classes with their corresponding text features by tuning both visual features and textual inputs. 
Thirdly, we train a conditional generative model using the well-aligned virtual image features and corresponding semantic text features, which enables us to generate labeled data for new classes. And then zero-shot classification is achieved through supervised learning. 
Our framework has been evaluated on five commonly used benchmarks for generalized ZSL, as well as 11 benchmarks for the base-to-new generalization. The results demonstrate the superiority of our approach.

\section{Related Work}
\textbf{Traditional Zero-Shot Learning.}
Zero-shot learning (ZSL) is a research area that explores the generalizability of deep learning models. Specifically, it focuses on training a classifier that can recognize samples from the new classes that are unseen during training. It is broadened to generalized zero-shot learning (GZSL) where both base and new classes should be recognized during the testing phase.
Embedding based methods and generative-model based methods are the two mainstream methodologies for GZSL.
An embedding based approach aims to learn a mapping function that maps visual features and semantic information into a unified space~\cite{t18,t19,t20,t21,t22}. The absence of new class data makes it prone to overfitting so that the test samples of the new classes are easy to be incorrectly classified into a base class.
To mitigate the data imbalance problem, recent studies prefer generative-model based methods because they can convert the challenging ZSL problem into a fully-supervised recognition task by synthesizing the absent samples of the new classes. 
Most generative-model based methods use GAN or VAE for generation~\cite{SR-GAN,t32,t33,DCRGAN}, and some studies have explored the combination of them which we termed as VAEGAN~\cite{t29,t30}. 
A significant drawback in both embedding based methods and generative-model based methods is that they rely on a large amount of real image data to learn the shared embedding space or train the generator. This requirement raises concerns related to copyright infringement and privacy issues. Moreover, these methods often necessitate experts attribute annotations, which are labor-intensive and expensive.

\textbf{Fine-Tuning for Vision-Language Models.}
Recently, the contrastive trained vision-language model CLIP~\cite{CLIP} shows impressive zero-shot performance on recognition tasks. When we directly infer with the pre-trained CLIP, which is denoted as Zero-Shot CLIP, the performance is still limited on some downstream datasets. It is because of the domain shift between the pre-training dataset and the downstream datasets for specific tasks, especially when the task is fine-grained.
Adapter methods focus on further learning a mapping network for the output features. One of them is CLIP-Adapter~\cite{t18}, which uses a residual connected MLP after the last vision layer and the last text layer.
Prompt tuning methods introduce parameters to the input. CoOp~\cite{CoOp} establishes a set of learnable vectors on the textual side to learn a generic prompt template. CoCoOp ~\cite{Co-CoOp} leverages information from the visual side and builds instance-level prompt templates to achieve base-to-new generalization. VP~\cite{VP} and VPT~\cite{vpt} design the tunable visual prompts either on the images or on the patch tokens.
MaPLe~\cite{maple} inserts learnable tokens inside the visual encoder and text encoder for deep fine-tuning, but it still relies on images and requires more training.
Unlike existing prompt tuning methods that focus on learning single-modality prompts or rely on the original images, our approach involves tuning prompts for both visual features (without need of the original images) and textual inputs. 


\textbf{Data-Free Transfer Learning.}
Data plays a significant role in the development of artificial intelligence. The value of data is more and more appreciative so real data is always not accessible due to copyright or privacy issues nowadays. Thus, data-free transfer learning, which just needs the source model for training but leaves the source data protected, receives increased attention.  Existing research under the Data-Free setting can be divided into three categories based on the level of the guards of the source model parameters: available source model parameters~\cite{t7,t43}, inaccessible parameters but enable gradient propogation~\cite{t44}, and a black-box service, where the source model just exposes an API for the client to request predictions~\cite{t7}. Despite their exhaustive experiments with various guard levels, the results are not satisfactory in the most realistic black-box scenario.

\begin{figure*}[htbp]
\centering
\includegraphics[width=0.95\linewidth,keepaspectratio]{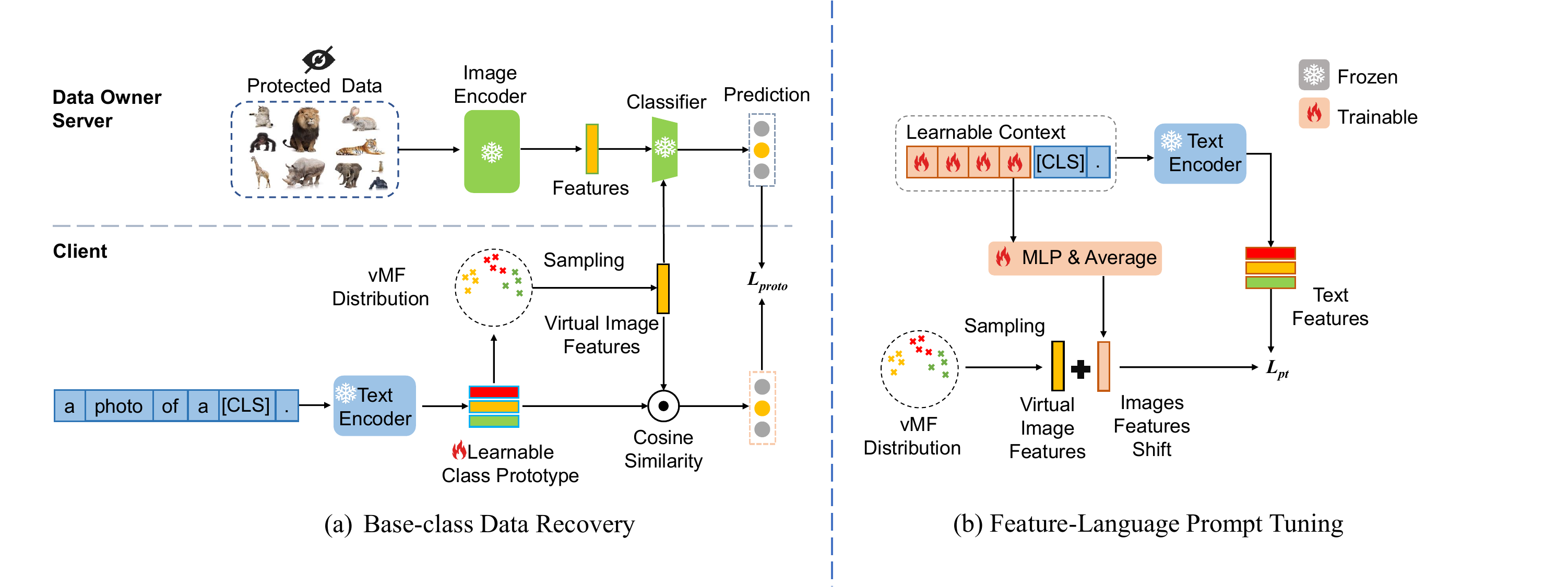}
\vspace{-1em}
\caption{The proposed framework is based on vision-language pre-trained models, such as CLIP. (a) Stage 1: Model the distribution of base class image features properly and then sample virtual image features. (b) Stage 2: Align the obtained virtual image features with the extracted text features via FLPT (Feature-Language Prompt Tuning). 
}
\label{figure1}
\vspace{-1em}
\end{figure*}

\section{Methodology}


\subsection{Problem Formulation and Overview}
In this paper, we propose and address a novel problem called data-free zero-shot learning (DFZSL). In the data-free setting, data from base classes are protected considering privacy or copyrights and cannot be directly used for training zero-shot learning methods. 

Formally, consider a data owner as the server and there are images of base classes on the server. The image encoder of pre-trained CLIP can be used to extract image features $ \bm{x}^{base} \in X^{base} $. The corresponding labels for base classes are $ y^{base} \in Y^{base} $. A classifier for base classes can be trained with image features extracted from real images by the pre-trained CLIP model. We denote the trained base classifier as $ f^{base}: X^{base} \to Y^{base} $. In a white-box scenario, the classifier weights are available. While in a black-box scenario, the server does not leak the classifier weights and just exposes an API for the client to request the prediction. 

The client side does not have access to any images that can be used for training in the data-free setting. The image features and labels of new class data are denoted as $ \bm{x}^{new} \in X^{new} $ and $ y^{new} \in Y^{new} $, respectively. The classes of base data and new data are disjoint so that $ Y^{base} \cap Y^{new} = \emptyset $. Now we have $ X = X^{base} \cup X^{new} $ and $ Y = Y^{base} \cup Y^{new} $. 

In data-free zero-shot learning, the objective is to classify test images from both base and new classes on the client side, utilizing the assistance of a base classifier located on the server, without directly accessing the training data.

\textbf{Overview.} To solve the proposed data-free zero-shot learning problem, we propose a generic framework consisting of three main stages. The overall pipeline is shown in Fig.~\ref{figure1}. Firstly, we propose to recover the image features of base classes from the base classifier. Secondly, we leverage the CLIP text features as the semantic side information, and propose a feature-language prompt tuning method to extract well-aligned high-quality image features and text features. Thirdly, with both, we follow traditional generative model-based methods to train the generator and classifier for the downstream fine-grained zero-shot classification tasks.

\subsection{Base-Class Data Recovery}
While we want to transfer from base classes to disjoint new classes on downstream datasets via traditional generative methods, we have to prepare the training data. Therefore, we propose a novel method to recover the image features of base classes from the base classifier located on the server in both white-box and black-box setting. The general idea is to model the distribution of the original data properly and then sample virtual data from it.

The most commonly used softmax classifier in deep learning is in the form of a matrix that consists of weights for each class and is trained with cross-entropy loss. During the training process, these weights are forced to be closer to the image features of the corresponding class and keep away from other classes. Especially for CLIP, cosine similarity is used as the metric for prediction. As a result, the L2-normed weights are good representations of class prototypes.  

To recover the base-class data, we hypothesise that the L2-normed image features follow a distribution on a hyper-sphere, where features belonging to the same class tend to cluster together. Therefore, we approximate the real image features of base classes using the von Mises-Fisher (vMF) distribution that is defined on the surface of a unit hyper-sphere. 
A vMF distribution is determined by two parameters, the mean direction $ \bm{\mu} $ and the concentration $ \kappa $. We establish a set of class prototypes: $ M = \left\{ \bm{\mu}_1, \bm{\mu}_2, \cdots, \bm{\mu}_{\left| Y^{base} \right|} \right\} $ as the mean direction of each class. In the white-box scenario, the base classifier weights $ W = \left\{ \bm{w}_1, \bm{w}_2, \cdots, \bm{w}_{\left| Y^{base} \right|} \right\} $ are available, and we directly set $ M = W $. In the black-box scenario, we turn the class prototypes $M$ into learnable parameters and initialize them by the text features $ T = \left\{ \bm{t}_1, \bm{t}_2, \cdots, \bm{t}_{\left| Y^{base} \right|} \right\} $ of corresponding classes. The text features are extracted by CLIP's text encoder $ \mathcal{T}_{encoder} $ from the prompt ``a photo of a [CLS].'', where [CLS] is a class name. 

With the mean direction, virtual image features can be sampled surrounding the center and the concentration controls the spread of sampled features. While we choose a proper concentration, the underlying principle is that if sampled features from the distribution belong to different classes, they should be sufficiently separable. 
We assume an isotropic covariance for simplicity. In statistics, vMF distribution is a close approximation to the wrapped Gaussian distribution.
According to the empirical rules of the Gaussian distribution, approximately 99.73\% of the sampled data deviates from the mean within a range of 3 standard deviations $ 3\sigma $. Therefore, if we aim for the sampled virtual data from the two classes to be discriminative, the arc length between their respective prototypes should be at least greater than $ 6\sigma $. Then the concentration parameter can be derived from the definition of the von Mises-Fisher (vMF) distribution, expressed as $ \kappa = \frac{1}{\sigma^2} $. Thus, we set the concentration that caters to all class pairs in the downstream datasets with the class prototypes: 
\begin{equation}\label{eqn-1}
\small
\hspace{-0.6em}
        \kappa_{text} = \max_{\forall y, y^{\prime} \in Y^{base}, y \ne y^{\prime}} \left\{ \left[ \frac{1}{6} \arccos \left( \frac{\bm{\mu}_y}{\left| \bm{\mu}_y \right|} \cdot \frac{\bm{\mu}_{y^{\prime}}}{\left| \bm{\mu}_{y^{\prime}} \right|} \right) \right]^{-2} \right\}
\end{equation}

The virtual image features can be considered to follow the vMF distribution determined by prototypes $M$, $ \lambda $, and $\kappa_{text}$:
\begin{equation}\label{eqn-2}
        \tilde{\bm{x}} \sim \mathrm{vMF}  \left(M, \lambda \cdot \kappa_{text} \right),
\end{equation}
where $ \lambda $ is a hyper-parameter to refine the concentration (which is nomally set to $\lambda=1 $ and a greater $\lambda$ produces a more concentrated distribution within each class).

We can now sample virtual image features of base classes $ \tilde{\bm{x}}^{base} \in \tilde{X}^{base} $ from the vMF distribution. However, for the black-box scenario, we further tune the initial class prototypes to mitigate the modality gap between image and text. We first upload the sampled virtual image features as test samples to the base classifier on the server and obtain the prediction scores $ s^{base} $. Subsequently, we treat the learnable class prototypes as a base classifier on the client and predict scores for these virtual image features. The objective is to align the two sets of scores and make the prototypes closely resemble the classifier weights protected on the server:
\begin{equation}\label{eqn-3}
\small
    \hspace{-0.5em}
        L_{proto}(M) = \frac{1}{\left| Y^{base} \right|}  \sum_{y \in Y^{base}} \left( \bm{s}^{base} - \cos \left( \tilde{\bm{x}}^{base}, M \right) \right)^2
\end{equation}

\subsection{Feature-Language Prompt Tuning}
In addition to the recovered virtual image features, another crucial component for achieving the transfer from base to new classes is the semantic information. 
We use the text features extracted by the pre-trained CLIP model. However, due to the discrepancy in class granularity between the pre-trained models and the downstream fine-grained zero-shot classification tasks at hand, the semantic information produced by the CLIP models trained on weakly aligned image-caption pairs is sub-optimal.
  
To further enhance the quality of the semantic information, we propose a feature-language prompt tuning method. On the visual side, our method requires only the features of the image, not the original picture. Considering that the recovered virtual image features only cover base classes and we have to transfer to new classes, we tune the prompts in a class-agnostic way.
Particularly, we replace the embeddings of class-agnostic prefix ``a photo of a'' used in CLIP with learnable parameters $ P=\left\{ \bm{p}_1, \bm{p}_2, \bm{p}_3, \bm{p}_4 \right\} $. In addition to the text prompts, we also introduce a shift term $ \bm{x}_{shift} $ as class-agnostic learnable parameters added to the image features. To accumulate class-agnostic generalizable knowledge, we establish a connection between the text prompts and the image shift using a light mapping network $ F_{\Theta} $ parameterized by $ \Theta $. It serves as a bridge between the textual information and the image transformation, allowing for the integration of both modalities. The $ \bm{x}_{shift} $ is then defined as:

\begin{equation}\label{eqn-4}
        \bm{x}_{shift} = \frac{1}{4} \sum_{i=1}^4 F_{\Theta} \left( P\right).
\end{equation}

We try to tune these parameters to better align the features of two modalities and mitigate the domain shift between CLIP's pre-training dataset and the downstream datasets. Then we have the enhanced image features which can help us easily generalize to different classes:

\begin{equation}\label{eqn-5}
\hat{\bm{x}}^{base} = \tilde{\bm{x}}^{base} + \alpha \bm{x}_{shift}, 
\end{equation} 
where $ \alpha $ is the trade-off parameter that controls how much we add the shift term. The enhanced text features are:
\begin{equation}\label{eqn-6}
\hat{\bm{t}} = \mathcal{T}_{encoder} \left( \left\{ \bm{p}_1\ \bm{p}_2\ \bm{p}_3\ \bm{p}_4\ \left[ CLS \right].  \right\} \right),
\end{equation}
where $ \left[ CLS \right] $ represents the embedding of class name. The parameters are optimized by lower the cross-entropy when classifying the image features by the text features:
\begin{equation}\label{eqn-7}
\small 
        L_{pt}(\Theta,P) = - \log \frac{\exp \left( { \cos \left( \hat{\bm{x}}^{base}, \hat{\bm{t}}_{y} \right)}/{\tau} \right)}{\sum_{y' \in Y^{base}} \exp \left( { \cos \left( \hat{\bm{x}}^{base}, \hat{\bm{t}}_{y'} \right)}/{\tau} \right)},
\end{equation}
and $ \tau $ is the temperature used in CLIP which equals to 0.01.

\subsection{New Class Features Generation for Zero-shot Classification}
After we recovered the base class image features and conducted the feature-language prompt tuning, we have already prepared high-quality training data for traditional generative zero-shot learning methods.

\textbf{Generate Data of New Classes.} We choose to train a suitable generative model based on the enhanced base data. The loss function is termed as follows:
\begin{equation}\label{eqn-8}
        L_{generator}(\Phi)=\ell \left( \hat{\bm{x}}^{base}, G_{\Phi} \left( \bm{z}, \hat{\bm{t}}^{base} \right) \right),
\end{equation}
where $\ell$ can be GAN loss or other loss defined by the chosen generative model. $G$ is the generator parameterized by $ \Phi $ and $ \bm{z} $ represents the random noise. 
Then we condition the generator with text features of new classes and generate the new class image features:
\begin{equation}\label{eqn-9}
        \hat{\bm{x}}^{new} = G_{\Phi} \left( \bm{z}, \hat{\bm{t}}^{new} \right).
\end{equation}

\textbf{Supervised Image Classification.} With the enhanced virtual image features of base classes $ \hat{\bm{x}}^{base} $ and the generated virtual image features of new classes $ \hat{\bm{x}}^{new} $, the generalized zero-shot learning problem is converted into a fully-supervised image classification problem. Moreover, we initialize the weights of the final classifier by the enhanced text features to speed up the training process.

\section{Experiments}
\subsection{Datasets and Implementation Details}
\textbf{Datasets.} We evaluate our method in two different tasks: generalized zero-shot learning and base-to-new generalization. 
For generalized zero-shot learning, we follow the same setting as ~\cite{s41:AWA2}. Our framework is evaluated on five datasets: Attribute Pascal and Yahoo (APY)~\cite{APY}, CaltechUCSD-Birds (CUB)~\cite{CUB}, Oxford Flowers  (FLO)~\cite{Flowers102}, SUN Attribute (SUN)~\cite{SUN}, and Animals with Attributes2 (AWA2)~\cite{s41:AWA2}, which contains 32, 200, 102, 717 and 50 classes, respectively.
As for the base-to-new generalization, we follow the setting proposed in CoCoOp ~\cite{Co-CoOp}. We evaluate the performance of our framework on 11 different image classification datasets which covers a wide range of recognition tasks.  This includes a large-scale visual dataset, ImageNet~\cite{imagenet}; a generic-objects datasets, Caltech101~\cite{Caltech101}; five fine-grained image recognition datasets, OxfordPets~\cite{OxfordPets}, StanfordCars~\cite{StanfordCars}, Flowers102~\cite{Flowers102}, Food101~\cite{food101} and FGVCAircraft~\cite{FGVCAircraft}; a satellite-view topographic image dataset EuroSAT~\cite{EuroSAT}; an action recognition dataset UCF101~\cite{UCF101}; a texture dataset DTD~\cite{DTD}; and a scene recognition dataset SUN397~\cite{SUN397}.
These datasets will be detailed in the appendix.

\textbf{Implementation Details.}
Our proposed framework consists of three stages: recover virtual image features of base classes, utilize FLPT to enhance both the virtual image features and semantic text features, and finally adopt traditional generative ZSL methods.
We recover the base class data in the data-free setting at the first stage. In the white-box scenario, where the base classifier weights are accessible, we directly apply them as the class prototypes. In the black-box scenario, the class prototypes are initialized by text features, and $ \lambda $ is set to 1. We apply the Adam optimizer and the learning rate is set to 0.0003.
In the second prompt tuning stage, we implement the light mapping network with a single-hidden-layer MLP activated by GELU.
For the third stage, we use off-the-shelf generative-model based methods. The setup stays the same as what they proposed in their papers.
All experiments are performed on an NVIDIA GeForce RTX3090, except for the ImageNet, which is performed on an NVIDIA A100.
\begin{table*}[]
\centering
\fontsize{15.5}{13}\selectfont
\resizebox{\linewidth}{!}{
\begin{tabular}{l|l|l|ccc|ccc|clc|ccc|ccc}
\hline
                            &                            &                & \multicolumn{3}{c|}{\textbf{AWA2}}                                                                & \multicolumn{3}{c|}{\textbf{APY}}                                                                          & \multicolumn{3}{c|}{\textbf{CUB}}                                                                    & \multicolumn{3}{c|}{\textbf{SUN}}                                                                 & \multicolumn{3}{c}{\textbf{FLO}}                                                                 \\
                            &                            &                & \multicolumn{1}{c}{\textbf{Base}} & \multicolumn{1}{c}{\textbf{New}} & \multicolumn{1}{c|}{\textbf{H}} & \textbf{Base}                        & \textbf{New}                        & \textbf{H}                         & \multicolumn{1}{c}{\textbf{Base}}    & \multicolumn{1}{c}{\textbf{New}} & \multicolumn{1}{c|}{\textbf{H}} & \multicolumn{1}{c}{\textbf{Base}} & \multicolumn{1}{c}{\textbf{New}} & \multicolumn{1}{c|}{\textbf{H}} & \multicolumn{1}{c}{\textbf{Base}} & \multicolumn{1}{c}{\textbf{New}} & \multicolumn{1}{c}{\textbf{H}} \\ \hline
\multirow{10}{*}{\rotatebox{90}{Resnet-101}} & \multirow{9}{*}{Real-Data} & HSVA           & 79.8                           & 56.7                           & 66.3                            & -                                 & -                                 & -                                  & 58.3                              & 52.7                           & 55.3                            & 48.6                           & 39.0                           & 43.3                            & \multicolumn{1}{c}{-}          & \multicolumn{1}{c}{-}          & \multicolumn{1}{c}{-}          \\
                            &                            & MSDN           & 74.5                           & 62.0                           & 67.7                            & -                                 & -                                 & -                                  & 67.5                              & \textbf{68.7}                  & 68.1                            & 34.2                           & 52.2                           & 41.3                            & \multicolumn{1}{c}{-}          & \multicolumn{1}{c}{-}          & \multicolumn{1}{c}{-}          \\
                            &                            & f-CLSWGAN      & 61.4                           & 57.9                           & 59.6                            & -                                 & -                                 & -                                  & 57.7                              & 3.7                            & 49.7                            & 36.6                           & 42.6                           & 39.4                            & 73.8                           & 59.0                           & 65.6                           \\
                            &                            & Cycle-WGAN     & 63.4                           & 59.6                           & 59.8                            & -                                 & -                                 & -                                  & 59.3                              & 47.9                           & 53.0                            & 33.8                           & 47.2                           & 39.4                            & 69.2                           & 61.6                           & 65.2                           \\
                            &                            & LisGAN         & 76.3                           & 52.6                           & 62.3                            & -                                 & -                                 & -                                  & 57.9                              & 46.5                           & 51.6                            & 37.8                           & 42.9                           & 40.2                            & 83.8                           & 57.7                           & 68.3                           \\
                            &                            & f-VAEGAN       & 70.6                           & 57.6                           & 63.5                            & -                                 & -                                 & -                                  & 60.1                              & 48.4                           & 53.6                            & 38.0                           & 45.1                           & 41.3                            & 74.9                           & 56.8                           & 64.6                           \\
                            &                            & TCN            & 65.8                           & 61.2                           & 63.4                            & \multicolumn{1}{l}{64.0}          & \multicolumn{1}{l}{24.1}          & \multicolumn{1}{l|}{35.1}          & 52.0                              & 52.6                           & 52.3                            & 37.3                           & 31.2                           & 34.0                            & \multicolumn{1}{c}{-}          & \multicolumn{1}{c}{-}          & \multicolumn{1}{c}{-}          \\
                            &                            & GCM-CF         & 75.1                           & 60.4                           & 67.0                            & \multicolumn{1}{l}{56.8}          & \multicolumn{1}{l}{37.1}          & \multicolumn{1}{l|}{44.9}          & 59.7                              & 61.0                           & 60.3                            & 37.8                           & 47.9                           & 42.2                            & \multicolumn{1}{c}{-}          & \multicolumn{1}{c}{-}          & \multicolumn{1}{c}{-}          \\
                            &                            & TF-VAEGAN      & 75.1                           & 59.8                           & 66.6                            & \multicolumn{1}{l}{61.5}          & \multicolumn{1}{l}{31.7}          & \multicolumn{1}{l|}{41.8}          & 64.7                              & 52.8                           & 58.1                            & 40.7                           & 45.6                           & 43.0                            & 84.1                           & 62.5                           & 71.7                           \\ \cline{2-18} 
                            & Data-Free                  & AZSL           & 3.7                            & 3.5                            & 3.6                             & \multicolumn{1}{l}{4.0}           & \multicolumn{1}{l}{6.8}           & \multicolumn{1}{l|}{5.1}           & \multicolumn{1}{c}{-}             & \multicolumn{1}{c}{-}          & \multicolumn{1}{c|}{-}          & \multicolumn{1}{c}{-}          & \multicolumn{1}{c}{-}          & \multicolumn{1}{c|}{-}          & \multicolumn{1}{c}{-}          & \multicolumn{1}{c}{-}          & \multicolumn{1}{c}{-}          \\
                            & Data-Free*                  & AZSL           & 44.3                            & 27.3                            & 33.7                             & \multicolumn{1}{l}{52.5}           & \multicolumn{1}{l}{17.9}           & \multicolumn{1}{l|}{26.7}           & \multicolumn{1}{c}{-}             & \multicolumn{1}{c}{-}          & \multicolumn{1}{c|}{-}          & \multicolumn{1}{c}{-}          & \multicolumn{1}{c}{-}          & \multicolumn{1}{c|}{-}          & \multicolumn{1}{c}{-}          & \multicolumn{1}{c}{-}          & \multicolumn{1}{c}{-}          \\ \hline
\multirow{7}{*}{\rotatebox{90}{CLIP}}      & \multirow{4}{*}{Real-Data} & f-VAEGAN       & 95.9                           & 61.2                           & 74.7                            & -                                 & -                                 & -                                  & \multicolumn{1}{c}{82.2} & \multicolumn{1}{c}{22.5}       & \multicolumn{1}{c|}{35.3}       & \multicolumn{1}{c}{-}          & \multicolumn{1}{c}{-}          & \multicolumn{1}{c|}{-}          & \textbf{97.6}                  & 11.1                           & 20.0                           \\
                            &                            & TF-VAEGAN      & \textbf{96.3}                  & 43.7                           & 60.1                            & 71.7                              & 22.3                              & 34.0                               & \multicolumn{1}{c}{\textbf{84.4}}         & \multicolumn{1}{c}{21.1}       & \multicolumn{1}{c|}{34.0}       & \multicolumn{1}{c}{51.4}       & \multicolumn{1}{c}{61.4}       & \multicolumn{1}{c|}{55.9}       & 97.2                           & 37.4                           & 54.0                           \\
                            &          & CoOp           & 95.3                           & 72.7                           & 82.5                            & \multicolumn{1}{l}{85.4}          & \multicolumn{1}{l}{76.1}          & \multicolumn{1}{l|}{80.5}          & 63.8                              & 49.2                           & 55.6                            & 61.3                           & 61.8                           & 61.6                            & 85.8                           & 52.2                           & 64.9                           \\
                            &                            & SHIP+CoOp      & 94.4                           & 84.1                           & 89.0                            & -                                 & -                                 & -                                  & \multicolumn{1}{c}{58.9}          & \multicolumn{1}{c}{55.3}       & \multicolumn{1}{c|}{57.1}       & \multicolumn{1}{c}{-}          & \multicolumn{1}{c}{-}          & \multicolumn{1}{c|}{-}          & 76.3                           & 69.0                           & 72.4                           \\ \cline{2-18} 
                            & Data-Free                 & CLIP*          & 93.0                          & 88.2                          & 90.6                           & 81.6                          & 75.8                          & 78.6                           & 56.3                          & 56.1                          & 56.2                          & 51.2                          & 55.9                          & 53.5                           & 69.4                          & 67.9                          & 69.6                          \\
                            & Data-Free                  & FLPT+TF-VAEGAN & 93.9                           & 93.2                           & 93.5                            & \multicolumn{1}{l}{84.2}          & \multicolumn{1}{l}{\textbf{81.1}} & \multicolumn{1}{l|}{82.6}          & 66.1                              & \textbf{60.9}                  & 63.4                            & 60.9                           & \textbf{65.6}                  & 63.2                            & 89.0                           & 78.2                           & 83.2                           \\
                            & Data-Free*                 & FLPT+TF-VAEGAN & 93.9                           & \textbf{93.6}                  & \textbf{93.7}                   & \multicolumn{1}{l}{\textbf{84.4}} & \multicolumn{1}{l}{81.0}          & \multicolumn{1}{l|}{\textbf{82.7}} & 70.4                              & 60.8                           & \textbf{65.2}                   & \textbf{63.8}                  & 62.8                           & \textbf{63.3}                   & 89.7                           & \textbf{79.5}                  & \textbf{84.3}                  \\ \hline
\end{tabular}}
\caption{{Generalized zero-shot learning.} 
The model is trained on the base classes and is evaluated on the \textbf{mixture} of base classes and new classes. `Base' indicates the base-class results, `New' indicates the new-class results, and `H' is the harmonic mean.
`Data-Free' represents the black-box scenario. `Data-Free*' means the white-box scenario. `CLIP*' means that the hand-crafted prompt templates are used.}
\label{table1}
\end{table*}

\subsection{Results of Generalized Zero-Shot Learning}
\textbf{Setup.} We follow the splits and evaluation protocols proposed in ~\cite{s41:AWA2}, train on base classes and then evaluated on test set that mixes the base classes and the new classes. Differently, the only input information of our base classes is the base classifier on the server. We do not use the attribute vectors provided in these benchmarks, but extract text features by CLIP with ``a photo of a [CLS].'' instead. The mean per-class top-1 accuracy is reported on base and new classes and the harmonic mean is computed to demonstrate the balanced performance of our framework.
\textbf{Baselines.} We choose f-CLSWGAN ~\cite{f-CLSWGAN}, Cycle-WGAN ~\cite{Cycle-WGAN}, LisGAN ~\cite{LisGAN}, TCN ~\cite{TCN}, f-VAEGAN ~\cite{f-VAEGAN}, TF-VAEGAN ~\cite{TF-VAEGAN}, GCMCF ~\cite{GCM-CF}, HSVA ~\cite{HSVA}  MSDN ~\cite{MSDN} AZSL~\cite{t7} and SHIP+CoOp~\cite{ship} as our baseline methods.

\textbf{Main results.} The results of generalized zero-shot learning on the five commonly used benchmarks are shown in Table \ref{table1}. Firstly, in comparison to traditional generalized zero-shot learning approaches that utilize the ImageNet-1k pre-trained ResNet-101 as the backbone, the baseline data-free ZSL method (AZSL~\cite{t7}) exhibits significantly inferior performance. This highlights the inherent challenge of the data-free ZSL task, as it performs notably worse than ZSL methods that have access to real base data. Secondly, our method improves both traditional generative ZSL methods using CLIP features and the state-of-the-art prompt-tuning methods even without access to the real data in both black-box (Data-Free) and white-box (Data-Free*) settings. For example, when compared with the state-of-the-art SHIP+CoOp method, our framework improves the harmonic mean accuracy by 4.5\%, 6.3\%, and 10.8\% on the three standard benchmarks of AWA2, CUB and FLO, respectively. This verify the effectiveness of the proposed framework. Thirdly, it is worth noting that the performance gain by our method does not solely originate from the base classes but primarily from the new classes. This observation validates the generalization ability of the proposed method.
\begin{table*}[]
\centering
\fontsize{10.5}{11}\selectfont
\resizebox{\linewidth}{!}{\begin{tabular}{l|l|lll|lll|lll|lll}
\hline
                           &                     & \multicolumn{3}{c|}{\textbf{11 Dataset Average}}                                                                  & \multicolumn{3}{c|}{\textbf{ImageNet}}                                                                 & \multicolumn{3}{c|}{\textbf{Caltech101}}                                                               & \multicolumn{3}{c}{\textbf{OxfordPets}}                                                               \\
                           &                     & \multicolumn{1}{c}{\textbf{Base}} & \multicolumn{1}{c}{\textbf{New}} & \multicolumn{1}{c|}{\textbf{H}} & \multicolumn{1}{c}{\textbf{Base}} & \multicolumn{1}{c}{\textbf{New}} & \multicolumn{1}{c|}{\textbf{H}} & \multicolumn{1}{c}{\textbf{Base}} & \multicolumn{1}{c}{\textbf{New}} & \multicolumn{1}{c|}{\textbf{H}} & \multicolumn{1}{c}{\textbf{Base}} & \multicolumn{1}{c}{\textbf{New}} & \multicolumn{1}{c}{\textbf{H}} \\ \hline
\multirow{7}{*}{Real-Data} & CoOp                & 82.69                             & 63.22                            & 71.66                           & 76.47                             & 67.88                            & 71.92                           & 98.00                             & 89.81                            & 93.73                           & 93.67                             & 95.29                            & 94.47                          \\
                           & CoCoOp              & 80.47                             & 71.69                            & 75.83                           & 75.98                             & 70.43                            & 73.10                           & 97.96                             & 93.81                            & 95.84                           & 95.20                             & 97.69                            & 96.43                          \\
                           & MaPLe               & 82.28                             & 75.14                            & 78.55                           & 76.66                             & 70.54                            & \textbf{73.47}                  & 97.74                             & 94.36                            & 96.02                           & 95.43                             & 97.76                            & 96.58                          \\
                           & CLIP-Adapter        & 83.05                             & 65.20                            & 73.05                           & 75.74                             & 68.21                            & 71.78                           & 98.13                             & 92.19                            & 95.39                           & 91.55                             & 90.10                            & 90.82                          \\
                           & CoOp + VPT          & 71.98                             & 74.76                            & 73.34                           & 74.73                             & \textbf{70.60}                   & 72.60                           & 95.47                             & 93.80                            & 94.62                           & 90.77                             & 97.83                            & 94.16                          \\
                           & SHIP + CoOp         & 80.03                             & 73.69                            & 76.73                           & 75.87                             & 69.95                            & 72.79                           & 97.55                             & 95.20                            & 96.36                           & 95.37                             & 97.87                            & 96.61                          \\
                           & SHIP + CLIP-Adapter & 83.14                             & 67.77                            & 74.67                           & 76.00                             & 69.32                            & 72.51                           & 97.68                             & 95.09                            & 96.37                           & 92.19                             & 93.85                            & 93.01                          \\ \hline
\multirow{3}{*}{Data-Free} & CLIP*                & 69.34                             & 74.22                            & 71.70                           & 72.43                             & 68.14                            & 70.22                           & 96.84                             & 94.00                            & 95.40                           & 91.17                             & 97.26                            & 94.12                          \\
& FLPT                & 78.08                             & 75.46                            & 76.85                           & 74.06                             & 68.74                            & 71.30                           & 97.87                             & \textbf{96.29}                   & 97.07                           & 95.59                             & 97.76                            & 96.66                          \\
                           & FLPT+TFVAEGAN       & \textbf{83.91}                    & \textbf{76.21}                   & \textbf{79.71}                  & \textbf{76.99}                    & 68.22                            & 72.34                           & \textbf{98.64}                    & 96.18                            & \textbf{97.40}                  & \textbf{96.49}                    & \textbf{98.21}                   & \textbf{97.34}                 \\ \hline  \hline

                           &                     & \multicolumn{3}{c|}{\textbf{StanfordCars}}                                                             & \multicolumn{3}{c|}{\textbf{Flowers102}}                                                               & \multicolumn{3}{c|}{\textbf{Food101}}                                                                  & \multicolumn{3}{c}{\textbf{FGVCAircraft}}                                                             \\
                           &                     & \multicolumn{1}{c}{\textbf{Base}} & \multicolumn{1}{c}{\textbf{New}} & \multicolumn{1}{c|}{\textbf{H}} & \multicolumn{1}{c}{\textbf{Base}} & \multicolumn{1}{c}{\textbf{New}} & \multicolumn{1}{c|}{\textbf{H}} & \multicolumn{1}{c}{\textbf{Base}} & \multicolumn{1}{c}{\textbf{New}} & \multicolumn{1}{c|}{\textbf{H}} & \multicolumn{1}{c}{\textbf{Base}} & \multicolumn{1}{c}{\textbf{New}} & \multicolumn{1}{c}{\textbf{H}} \\ \hline
\multirow{7}{*}{Real-Data}
                           & CoOp                & 78.12                             & 60.40                            & 68.13                           & 97.60                             & 59.67                            & 74.06                           & 88.33                             & 82.26                            & 85.19                           & 40.44                             & 22.30                            & 28.75                          \\
                           & CoCoOp              & 70.49                             & 73.59                            & 72.01                           & 94.87                             & 71.75                            & 81.71                           & 90.70                             & 91.29                            & 90.99                           & 33.41                             & 23.71                            & 27.74                          \\
                           & MaPLe               & 72.94                             & 74.00                            & 73.47                           & 95.92                             & 72.46                            & 82.56                           & 90.71                             & 92.05                            & 91.38                           & 37.44                             & 35.61                            & 36.50                          \\
                           & CLIP-Adapter        & \textbf{79.16}                    & 59.49                            & 67.93                           & \textbf{98.29}                    & 64.68                            & 78.02                           & 88.24                             & 88.33                            & 88.29                           & 42.14                             & 25.67                            & 31.91                          \\
                           & CoOp + VPT          & 65.27                             & \textbf{75.97}                   & 70.21                           & 72.97                             & 75.90                            & 74.40                           & 90.37                             & 91.67                            & 91.01                           & 29.57                             & 33.80                            & 31.54                          \\
                           & SHIP + CoOp         & 68.57                             & 73.90                            & 71.14                           & 94.02                             & 74.40                            & 83.06                           & 90.54                             & 91.03                            & 90.78                           & 34.27                             & 32.33                            & 33.28                          \\
                           & SHIP + CLIP-Adapter & 78.51                             & 62.52                            & 69.61                           & 98.20                             & 65.89                            & 78.86                           & 88.63                             & 87.07                            & 87.84                           & 42.26                             & 30.05                            & 35.13                          \\ \hline

\multirow{3}{*}{Data-Free} & CLIP*                & 63.37                             & 74.89                            & 68.65                           & 72.08                             & 77.80                            & 74.83                           & 90.10                             & 91.22                            & 90.66                           & 27.19                             & \textbf{36.29}                   & 31.09                          \\
& FLPT                & 65.24                             & 75.74                            & 70.10                           & 88.79                             & 76.52                            & 82.20                           & 90.74                             & 92.09                            & 91.41                           & 32.89                             & 33.17                            & 33.03                          \\
                           & FLPT+TFVAEGAN       & 77.06                             & 75.41                            & \textbf{76.23}                  & 94.11                             & \textbf{78.65}                   & \textbf{85.69}                  & \textbf{91.71}                    & \textbf{92.10}                   & \textbf{91.90}                  & \textbf{45.26}                    & 32.81                            & \textbf{38.04}                
\\ \hline  \hline

                           &                     & \multicolumn{3}{c|}{\textbf{SUN397}}                                                                   & \multicolumn{3}{c|}{\textbf{DTD}}                                                                      & \multicolumn{3}{c|}{\textbf{EuroSAT}}                                                                  & \multicolumn{3}{c}{\textbf{UCF101}}                                                                   \\
                           &                     & \multicolumn{1}{c}{\textbf{Base}} & \multicolumn{1}{c}{\textbf{New}} & \multicolumn{1}{c|}{\textbf{H}} & \multicolumn{1}{c}{\textbf{Base}} & \multicolumn{1}{c}{\textbf{New}} & \multicolumn{1}{c|}{\textbf{H}} & \multicolumn{1}{c}{\textbf{Base}} & \multicolumn{1}{c}{\textbf{New}} & \multicolumn{1}{c|}{\textbf{H}} & \multicolumn{1}{c}{\textbf{Base}} & \multicolumn{1}{c}{\textbf{New}} & \multicolumn{1}{c}{\textbf{H}} \\ \hline
\multirow{8}{*}{Real-Data}
                           & CoOp                & 80.60                             & 65.89                            & 72.51                           & 79.44                             & 41.18                            & 54.24                           & 92.19                             & 54.74                            & 68.69                           & 84.69                             & 56.05                            & 67.46                          \\
                           & CoCoOp              & 79.74                             & 76.86                            & 78.27                           & 77.01                             & 56.00                            & 64.85                           & 87.49                             & 60.04                            & 71.21                           & 82.33                             & 73.45                            & 77.64                          \\
                           & MaPLe               & 80.82                             & \textbf{78.70}                            & \textbf{79.75}                  & 80.36                             & 59.18                            & 68.16                           & 94.07                             & 73.23                            & 82.35                           & 83.00                             & \textbf{78.66}                   & \textbf{80.77}                 \\
                           & CLIP-Adapter        & 79.44                             & 66.81                            & 72.58                           & \textbf{81.94}                    & 39.49                            & 53.30                           & 93.45                             & 54.41                            & 68.78                           & 85.42                             & 67.77                            & 75.58                          \\
                           & CoOp + VPT          & 73.77                             & 77.90                   & 75.77                           & 57.67                             & 58.70                            & 58.18                           & 67.97                             & 71.63                            & 69.75                           & 73.23                             & 74.63                            & 73.92                          \\
                           & SHIP + CoOp        & 79.54                             & 75.27                            & 77.35                           & 74.88                             & 56.88                            & 64.65                           & 88.62                             & 66.87                            & 76.22                           & 81.08                             & 76.85                            & 78.91                          \\
                           & SHIP + CLIP-Adapter & 79.86                             & 66.52                            & 72.58                           & 81.60                             & 46.38                            & 59.14                           & 93.05                             & 57.15                            & 70.81                           & \textbf{86.61}                    & 71.61                            & 78.40                          \\ \hline
\multirow{3}{*}{Data-Free} & CLIP*                & 69.36                             & 75.35                            & 72.23                           & 53.24                             & 59.90                            & 56.37                           & 56.48                             & 64.05                            & 60.03                           & 70.53                             & 77.50                            & 73.85                          \\
& FLPT                & 77.84                             & 76.25                            & 77.04                           & 70.37                             & 62.68                            & 66.30                           & 86.00                             & 79.54                            & 82.64                           & 79.52                             & 75.66                            & 77.55                          \\
                           & FLPT+TF-VAEGAN       & \textbf{82.23}                    & 76.23                            & 79.12                           & 80.32                             & \textbf{64.37}                   & \textbf{71.47}                  & \textbf{95.74}                    & \textbf{79.97}                   & \textbf{87.15}                  & 84.50                             & 76.21                            & 80.13                          \\ \hline
\end{tabular}}
\caption{{Base-to-new generalization. }
The model is trained on the base classes and is evaluated on the base classes and new classes independently. `Base' indicates the base-class results, `New' indicates the new-class results, and `H' is the harmonic mean. `Data-Free' represents the black-box scenario. `Data-Free*' means the white-box scenario. `CLIP*' means that the hand-crafted prompt templates are used.}
\label{table2}
\end{table*}

\begin{table*}[]
\centering
\fontsize{22}{25}\selectfont
\resizebox{\linewidth}{!}{
\begin{tabular}{l|l|lll|lll|lll|lll|lll}
\hline
                           &                & \multicolumn{3}{c|}{\textbf{AWA2}}                                                                & \multicolumn{3}{c|}{\textbf{APY}}                                                                 & \multicolumn{3}{c|}{\textbf{CUB}}                                                                 & \multicolumn{3}{c|}{\textbf{SUN}}                                                                 & \multicolumn{3}{c}{\textbf{FLO}}                                                                 \\
                           &                & \multicolumn{1}{c}{\textbf{Base}} & \multicolumn{1}{c}{\textbf{New}} & \multicolumn{1}{c|}{\textbf{H}} & \multicolumn{1}{c}{\textbf{Base}} & \multicolumn{1}{c}{\textbf{New}} & \multicolumn{1}{c|}{\textbf{H}} & \multicolumn{1}{c}{\textbf{Base}} & \multicolumn{1}{c}{\textbf{New}} & \multicolumn{1}{c|}{\textbf{H}} & \multicolumn{1}{c}{\textbf{Base}} & \multicolumn{1}{c}{\textbf{New}} & \multicolumn{1}{c|}{\textbf{H}} & \multicolumn{1}{c}{\textbf{Base}} & \multicolumn{1}{c}{\textbf{New}} & \multicolumn{1}{c}{\textbf{H}} \\ \hline
\multirow{3}{*}{Data-Free} & CLIP*          & 93.04                          & 88.21                          & 90.57                           & 81.63                          & 75.76                          & 78.58                           & 56.29                          & 56.12                          & 56.21                           & 51.20                          & 55.90                          & 53.45                           & 69.39                          & 67.86                          & 69.62                          \\
                           & CoOp           & 94.35                          & 85.27                          & 89.58                           & 82.21                          & 78.47                          & 80.30                           & 51.90                          & 51.73                          & 51.82                           & 55.78                          & 53.19                          & 54.45                           & 76.82                          & 65.30                          & 70.59                          \\
                           & FLPT           & 93.57                          & 92.92                          & 93.24                           & 83.00                          & 80.11                          & 81.82                           & 57.26                          & 57.75                          & 57.50                           & 56.94                          & 59.72                          & 58.30                           & 70.54                          & 72.64                          & 71.57                          \\ \hline
Real-Data                  & FLPT           & 93.84                          & 93.60                          & 93.72                           & 84.50                          & 79.95                          & 82.16                           & 62.22                          & 59.44                          & 60.80                           & 57.64                          & 64.10                          & 60.69                           & 71.61                          & 73.20                          & 72.40                          \\ \hline \hline
\multirow{3}{*}{Data-Free} & CLIP*+ZLAP      & 94.05                          & 92.79                          & 93.42                           & 81.97                          & 76.90                          & 79.36                           & 63.13                          & 61.52                          & 62.31                           & 67.52                          & 51.39                          & 58.36                           & 87.62                          & 68.39                          & 76.82                          \\
                           & CoOp+ZLAP      & 94.50                          & 88.76                          & 91.54                           & 83.11                          & 79.69                          & 81.36                           & 64.43                          & 60.76                          & 62.54                           & 66.09                          & 52.85                          & 58.73                           & 89.64                          & 68.57                          & 77.70                          \\
                           & FLPT+ZLAP      & 94.37                          & 93.69                          & 94.03                           & 83.99                          & 80.59                          & 82.25                           & 65.80                          & 60.70                          & 63.14                           & 65.89                          & 62.29                          & 64.04                           & 91.53                          & 77.28                          & 83.80                          \\ \hline
Real-Data                  & FLPT+ZLAP      & 92.66                          & 96.05                          & 94.32                           & 83.54                          & 81.91                          & 82.72                           & 66.15                          & 62.80                          & 64.43                           & 59.53                          & 79.38                          & 68.04                           & 96.51                          & 74.27                          & 83.94                          \\ \hline \hline
\multirow{3}{*}{Data-Free} & CLIP*+SDGZSL    & 93.14                          & 93.69                          & 93.41                           & 82.02                          & 76.37                          & 79.09                           & 59.14                          & 54.51                          & 56.73                           & 58.10                          & 59.79                          & 58.93                           & 94.73                          & 62.68                          & 75.45                          \\
                           & CoOp+SDGZSL    & 93.43                          & 92.15                          & 92.79                           & 82.14                          & 79.31                          & 80.70                           & 57.04                          & 53.20                          & 55.05                           & 58.99                          & 61.18                          & 60.07                           & 94.37                          & 66.23                          & 77.83                          \\
                           & FLPT+SDGZSL    & 92.47                          & 95.30                          & 93.87                           & 84.10                          & 80.20                          & 82.10                           & 63.18                          & 65.38                          & 64.26                           & 59.50                          & 68.26                          & 63.58                           & 92.40                          & 74.12                          & 82.26                          \\ \hline
Real-Data                  & FLPT+SDGZSL    & 93.94                          & 93.03                          & 93.48                           & 83.72                          & 81.20                          & 82.44                           & 74.46                          & 62.09                          & 67.72                           & 64.11                          & 76.32                          & 69.68                           & 96.34                          & 74.17                          & 83.81                          \\ \hline \hline
\multirow{3}{*}{Data-Free} & CLIP*+TF-VAEGAN & 93.88                          & 89.23                          & 91.49                           & 81.93                          & 78.22                          & 80.04                           & 65.15                          & 53.14                          & 58.54                           & 63.06                          & 62.85                          & 62.95                           & 90.61                          & 65.96                          & 76.34                          \\
                           & CoOp+TF-VAEGAN & 94.51                          & 88.50                          & 91.41                           & 83.01                          & 79.39                          & 81.16                           & 63.04                          & 56.72                          & 59.71                           & 60.31                          & 61.46                          & 60.88                           & 89.63                          & 67.81                          & 77.21                          \\
                           & FLPT+TF-VAEGAN & 93.86                          & 93.16                          & 93.51                           & 84.18                          & 81.12                          & 82.62                           & 66.06                          & 60.92                          & 63.39                           & 60.93                          & 65.56                          & 63.16                           & 88.98                          & 78.15                          & 83.22                          \\ \hline
Real-Data                  & FLPT+TF-VAEGAN & 93.84                          & 93.59                          & 93.71                           & 84.56                          & 80.75                          & 82.61                           & 73.25                          & 58.45                          & 65.02                           & 62.87                          & 70.97                          & 66.67                           & 94.11                          & 77.58                          & 85.05                          \\ \hline
\end{tabular}}
\caption{{Ablation Study. } Comparison results of different prompt-tuning methods, different generative models, and `Data-Free' v.s. `Real-Data' settings. `CLIP*' means the hand-crafted prompt templates are used.}
\label{table3}
\vspace{-1em}
\end{table*}

\subsection{Results of Base-to-New Generalization}                 
\textbf{Setup.} We follow CoCoOp~\cite{Co-CoOp} to makes a half-and-half split on 11 datasets, which turns out to divide them into two non-overlapping subsets: the base classes and the new classes. The base-to-new generalization task requires the model to train on base classes and then separately
test on base classes and new classes. It is just the same as the conventional ZSL when the model is tested on new classes.

\textbf{Baselines.} We take several recent prompt-tuning methods as baselines, including CoOp, CoCoOp, MaPLe~\cite{maple}, CLIP-Adapter~\cite{CLIP-Adapter}, VPT~\cite{vpt} and SHIP~\cite{ship}. 

\textbf{Main Results.} As shown in Table \ref{table2}, most of the prompt-tuning methods improve the performance of CLIP in the base classes, while they demonstrate limited performance gain or even degradation for new classes. This may be due to the fact that training only on the base classes leads the model to overfit the base classes. 
To mitigate overfitting to the base classes, SHIP generates the new-class data via the pre-trained CLIP encoders. However, the CLIP features may not be optimal due to the domain gap between the pre-training data and the downstream task data, leading to sub-optimal results. By contrast, our method generates new-class data based on the proposed FLTP method that further align the visual and textual features through multi-modal prompt-tuning strategy. The promising results achieved by FLPT+TFVAEGAN show that the further aligned features are better for training an effective conditional generator.

\begin{figure}[htbp]
\vspace{1em}
\centering
\includegraphics[width=\linewidth]{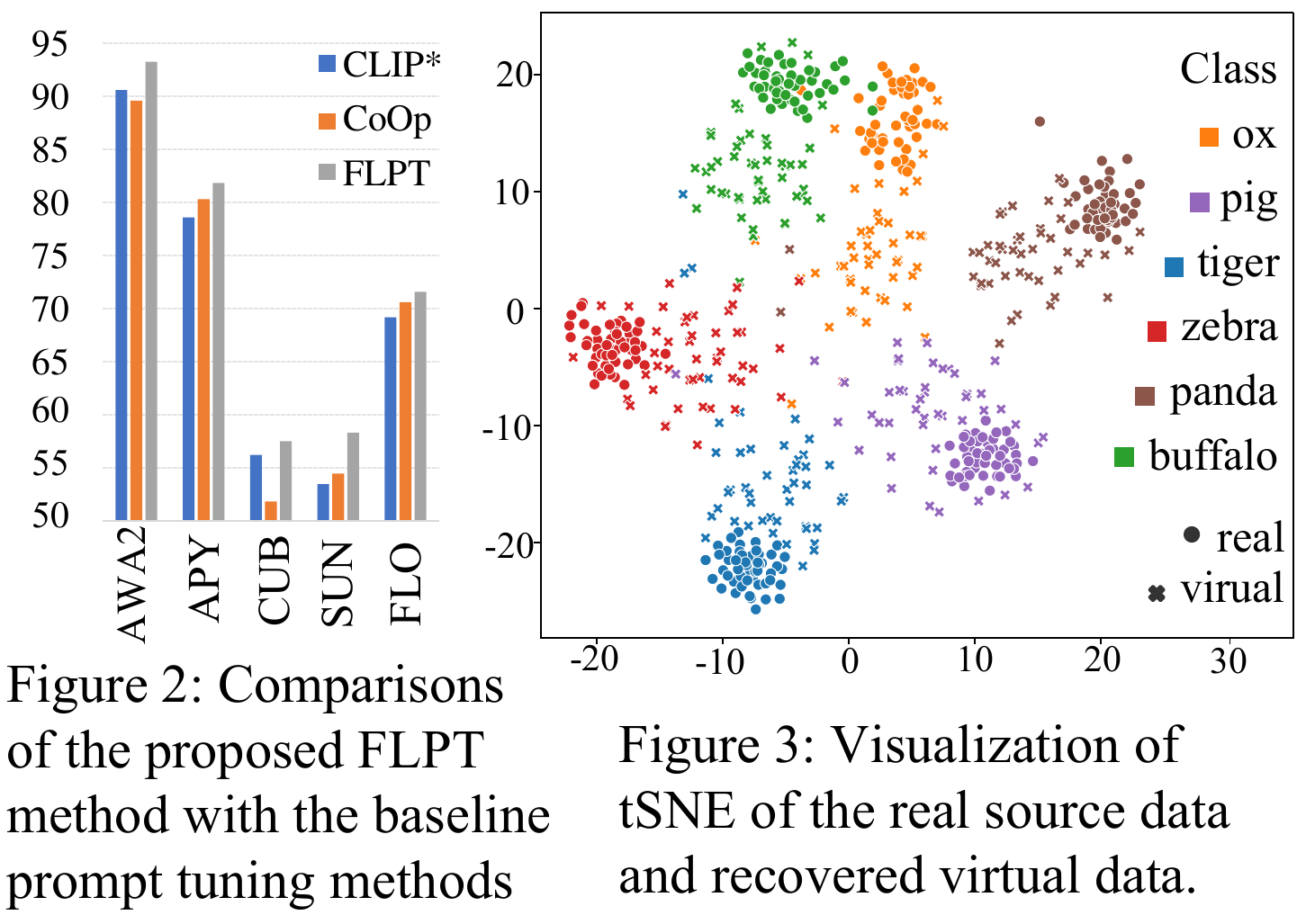}
\label{figure23}
\vspace{-3em}
\end{figure}

\subsection{Ablation Study}
We conducted ablation studies under the setting of generalized zero-shot learning to demonstrate the validity and generalizability of our proposed framework.

\textbf{Quality of the Recovered Virtual Data.}
We visualized the recovered virtual base data and the real base data of 6 classes in AWA2 in Figure 3. It can be seen that the recovered data exhibits a distribution similar to that of the real data and possesses sufficient class discriminative qualities. 
To further validate the quality of the recovered image features and make a fairer comparison with baselines, we performed the experiments on real and virtual data, respectively. 
As shown in Table~\ref{table3}, It can be seen that the performance gap between the real data and virtual data is small, especially for AWA2, APY and FLO, which validate the effectiveness of the proposed base class data recovery method. The gap is slightly larger for CUB and SUN datasets, which is due to their fine-grained and challenging nature.

\textbf{Comparisons of Different Prompt-tuning Methods.}
To evaluate the proposed FLPT that enhances the image features and text features, we compare FLPT to hand-crafted prompts in CLIP and the learning-based CoOp in Figure 2 and Table \ref{table3}. It can be seen that FLPT outperforms the other two on all the five GZSL benchmarks. Furthermore, the performance is further improved after applying the enhanced features to the generative-model based methods. Compared to hand-crafted prompts, FLPT is a data-driven method, which costs less and is more effective. While compared to CoOp, FLPT makes a link between two modalities and aligns both image features and text features simultaneously.

\textbf{Different Generative-Model-Based Methods.}
The proposed FLPT method can be combined with any generative models to further improve the ZSL performance. We evaluate our method with three generative methods: GAN-based ZLAP~\cite{t34:ZLAP}, VAE-based SDGZSL~\cite{SDGZSL}, and VAEGAN-based TF-VAEGAN~\cite{TF-VAEGAN}. 
As shown in Table \ref{table3}, when integrating FLPT with the three types of generative models, all of them exhibit improved performance compared to FLPT alone.
\section{Conclusion}
This paper addresses a challenging and practical problem dubbed as data-free zero-shot learning (DFZSL). In DFZSL, the use of real images from both the base classes and the new classes is not necessary, thereby effectively preserving data copyright and privacy. To tackle DFZSL, we propose a CLIP-based framework, which consists of three main stages. Firstly, the virtual base-class data are recovered via a modeled von Mises-Fisher distribution based on the pre-trained CLIP classifier. Secondly, we propose a feature-language prompt tuning method to further align the virtual image features and textual features. Thirdly, to achieve better zero-shot classification, we generate the new-class data by training a conditional generative model based on the well aligned base-class multi-modal features. Extensive experiments on both base-to-new ZSL and generalized ZSL demonstrate the effectiveness of the proposed framework.

\section*{Acknowledgments}
This work was supported by the National Key Research and Development Program of China (No. 2021YFB1714300), National Natural Science Foundation of China (No.62006012, No.62132001, No.62002012), in part by the Hong Kong Research Grants Council General Research Fund (17203023), in part by The Hong Kong Jockey Club Charities Trust under Grant 2022-0174,  in part by the Startup Funding and the Seed Funding for Basic Research for New Staff from The University of Hong Kong, and in part by the funding from UBTECH Robotics.
\bigskip

\bibliography{Data-Free_Generalized_Zero-Shot_Learning}

\end{document}